\documentclass{article}
\usepackage{ijcai21}

\usepackage{times}

\usepackage{soul}
\usepackage{url}
\usepackage[hidelinks]{hyperref}
\usepackage[utf8]{inputenc}
\usepackage[small]{caption}
\usepackage{graphicx}
\usepackage{amsmath}
\usepackage{booktabs}
\usepackage{pdfpages}
\usepackage{paralist}

\renewcommand{\texttt}[1]{\textit{#1}}

\usepackage{xcolor}  
\usepackage{tikz}  
\usepackage{amsmath}
\usepackage{amsthm,amssymb,amsthm}
\usepackage[ruled,vlined,linesnumbered]{algorithm2e}
\usepackage{lscape}
\usepackage{multirow}
\usepackage{hyperref}

\usepackage{listings} 
\SetKwComment{Comment}{$\triangleright$\ }{}

\DeclareMathOperator*{\argmin}{arg\,min}

\definecolor{ao(english)}{rgb}{0.0, 0.5, 0.0}
\newcommand{\NP}[1]{{\color{red}[NP: #1]}}
\newcommand{\ELM}[1]{{\color{blue}[ELM: #1]}}
\newcommand{\AZ}[1]{{\color{teal}[AZ: #1]}}
\newcommand{\MK}[1]{{\color{orange}[MK: #1]}}
\newcommand{\RM}[1]{{\color{ao(english)}[RM: #1]}}

\usepackage{bbm}
\usepackage{dsfont}

\newcommand\RR {{\mathds{R}}} 
\newtheorem{definition}{Def.}  

\newcommand\E {{\hat{y}}} 
\newcommand\C {{\mathcal{C}}} 
\newcommand\Net {{\mathcal{N}}} 

\newtheorem{proposition}{Prop.}

\title{On Guaranteed Optimal Robust Explanations for NLP Models}

\author{
 Emanuele La Malfa$^1$\footnote{First Author, contact at emanuele.la.malfa@cs.ox.ac.uk}\and
 Agnieszka Zbrzezny$^{1,2}$\and
 Rhiannon Michelmore$^{1}$\\
 Nicola Paoletti$^3$\And
 Marta Kwiatkowska$^1$\\
 \affiliations
 $^1$University of Oxford, $^2$University of Warmia and Mazury in Olsztyn\and \\ $^3$Royal Holloway, University of London\\
}

\begin{document}

\maketitle

\begin{abstract}
We build on abduction-based explanations for machine learning and develop a method for computing local explanations for neural network models in natural language processing (NLP). Our explanations comprise a subset of the words of the input text that satisfies two key features: optimality w.r.t. a user-defined cost function, such as the length of explanation, and robustness, in that they ensure prediction invariance for any bounded perturbation in the embedding space of the left-out words. We present two solution algorithms, respectively based on implicit hitting sets and maximum universal subsets, introducing a number of algorithmic improvements to speed up convergence of hard instances. We show how our method can be configured with different perturbation sets in the embedded space and used to detect bias in predictions by enforcing include/exclude constraints on biased terms, as well as to enhance existing heuristic-based NLP explanation frameworks such as Anchors. We evaluate our framework on three widely used sentiment analysis tasks and texts of up to $100$ words from SST, Twitter and IMDB datasets, demonstrating the effectiveness of the derived explanations\footnote{Code available at \url{https://github.com/EmanueleLM/OREs}}.

\end{abstract}

\section{Introduction}
The increasing prevalence of deep learning models in real-world decision-making systems has made AI explainability a central problem, as we seek to complement such highly-accurate but opaque models with comprehensible explanations as to 
why the model produced a particular prediction~\cite{samek2017explainable,ribeiro2016should,zhang2019interpreting,liu2018improving,letham2015interpretable}.
Amongst existing techniques, \textit{local explanations} explain the individual prediction in terms of a subset of the input features that justify the prediction. 
State-of-the-art explainers such as LIME and Anchors~\cite{ribeiro2016should,ribeiro2018anchors} use heuristics to obtain short explanations, which may generalise better beyond the given input and are more easily interpretable to human experts, but lack robustness to adversarial perturbations.  
The abduction-based method of~\cite{IgnatievNM19}, on the other hand, ensures minimality and robustness of the prediction by requiring its invariance w.r.t.\ any perturbation of the left-out features, meaning that the explanation is sufficient to imply the prediction. However, since perturbations are potentially unbounded, this notion of robustness may not be appropriate for certain applications.

In this paper, we focus on natural language processing (NLP) neural network models and, working in the embedding space with words as features, introduce \textit{optimal robust explanations (OREs)}. OREs are \textit{provably guaranteed} to be both \textit{robust}, in the sense that the prediction is invariant for any (reasonable) replacement of the features outside the explanation, and \textit{minimal} for a given user defined cost function, such as the length of the explanation. Our core idea shares similarities with abduction-based explanations (ABE) of~\cite{IgnatievNM19}, but is better suited to NLP models, where the unbounded nature of ABE perturbations may result in trivial explanations equal to the entire input. We show that OREs can be formulated as a particular kind of ABE or, equivalently, minimal satisfying assignment (MSA). We develop two methods to compute OREs by extending existing algorithms for ABEs and MSAs~\cite{IgnatievNM19,DilligDMA12}. In particular, we incorporate state-of-the-art robustness verification methods~\cite{KatzHIJLLSTWZDK19,wang2018efficient} to solve entailment/robustness queries and improve convergence by including sparse adversarial attacks and search tree reductions.  
By adding suitable constraints, we show that our approach allows one to detect biased decisions~\cite{darwiche2020reasons} and enhance heuristic explainers with robustness guarantees~\cite{ignatiev2019validating}. 

To the best of our knowledge, this is the first method to derive local explanations for NLP models with provable robustness and optimality guarantees. We empirically demonstrate that our approach can provide useful explanations for non-trivial fully-connected and convolutional networks on three widely used sentiment analysis benchmarks (SST, Twitter and IMDB).  
We compare OREs with the popular Anchors method, showing that Anchors often lack prediction robustness in our benchmarks, 
and demonstrate the usefulness of our framework on model debugging, bias evaluation, and repair of non-formal explainers like Anchors.

\section{Related Work}
\label{sec:rel_work}

\noindent Interpretability of machine learning models is receiving increasing attention~\cite{chakraborty2017interpretability}. Existing methods broadly fall in two categories: explanations via globally interpretable models (e.g. \cite{wang2015falling,zhang2018unsupervised}), and local explanations for a given input and prediction (to which our work belongs). Two prominent examples of the latter category are LIME~\cite{ribeiro2016should}, which learns a linear model around the neighbourhood of an input using random local perturbations, and Anchors~\cite{ribeiro2018anchors} (introduced in Section 3). These methods, however, do not consider robustness, making them fragile to adversarial attacks and thus insufficient to imply the prediction. Repair of non-formal explainers has been studied in~\cite{ignatiev2019validating} but only for boosted trees predictors.
\cite{narodytska2019assessing} assesses the quality of Anchors' explanations by encoding the model and explanation as a propositional formula. The explanation quality is then determined using model counting, but for binarised neural networks only. Other works that focus on binarised neural networks, Boolean classifiers or similar representations include \cite{shi2020tractable,darwiche2020reasons,darwiche2020three}. Methods 
tailored to (locally) 
explaining NLP model decisions for a given input include~\cite{li2015visualizing,singh2018hierarchical}. These identify input features, or clusters of input features, that most contribute to the prediction, using saliency and agglomerative contextual decomposition respectively. Layer-wise relevance propagation~\cite{bach2015pixel} is also popular for NLP explanations, and is used in~\cite{arras2016explaining,arras2017explaining,ding2017visualizing}. Similarly to the above, these methods do not consider robustness. Robustness of neural network NLP models to adversarial examples has been studied in~\cite{huang2019achieving,jia2019certified,la2020assessing}. We note that robustness verification is a different (and arguably simpler) problem from deriving a robust explanation, as the latter requires performing multiple robustness verification queries (see Section 4). 
Existing neural network verification approaches include symbolic (SMT)~\cite{KatzHIJLLSTWZDK19}, relaxation \cite{ko2019popqorn,wang2018efficient}, and global optimisation \cite{ruan2018reachability}. Research utilising hitting sets can be seen in \cite{IgnatievNM19-adv}, which relates explanations and adversarial examples through a generalised form of hitting set duality, and \cite{ignatiev2019model}, which works on improving model-based diagnoses by using an algorithm based on hitting sets to filter out non-subset-minimal sets of diagnoses.

\section{Optimal Robust Explanations for NLP}
\begin{figure}
\centering
\includegraphics[height=4.0cm,width=5.0cm]{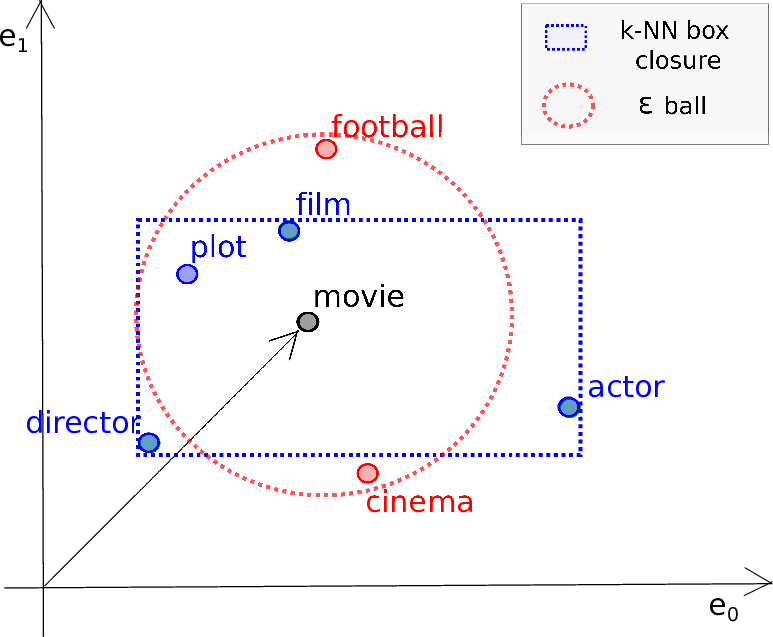}\caption{A graphical representation of the perturbation sets we define in the embedding space.}
\label{fig:knn-knn-box}
\end{figure}

\noindent\textbf{Preliminaries} \label{sec:problem-formlation}
We consider a standard NLP classification task where we classify some given input text $t$ into a plausible class $y$ from a finite set $\mathcal{Y}$. 
We assume that $t$ is a fixed length sequence of words (i.e., \textit{features})  $l$, $t=(w_1, \ldots, w_l)$, where $w_i\in W$ with $W$ being a finite vocabulary (possibly including padding). 
Text inputs are encoded using 
a continuous \textit{word embedding} $\mathcal{E}: W \rightarrow \mathbb{R}^d$, where $d$ is the size of the embedding~\cite{abs-1301-3781}. Thus, given a text $t=(w_1, \ldots, w_l)$, we define the embedding $\mathcal{E}(t)$ of $t$ as the sequence $x = (x_{w_1}, \ldots, x_{w_l}) \in \mathbb{R}^{l\cdot d}$, where $x_{w_i} = \mathcal{E}(w_i)$. We denote with $W_{\mathcal{E}} \subseteq W$ the vocabulary used to train $\mathcal{E}$. We consider embedding vectors trained from scratch 
on the sentiment task, a technique that enforces words that are positively correlated to each of the output classes to be gathered closer in the embedding space~\cite{baroni2014don}, 
which is considered a good proxy for semantic similarity with respect to the target task compared to count-based embeddings~\cite{alzantot2018generating}. 
For classification we consider a \textit{neural network} $M:\mathbb{R}^{l\cdot d} \rightarrow \mathcal{Y}$ that operates on the text embedding. 
\\
\noindent\textbf{Robust Explanations} 
In this paper, we seek to provide \textit{local explanations} for the predictions of a neural network NLP model. For a text embedding $x=\mathcal{E}(t)$ and a prediction $M(x)$, a local explanation $E$ is a subset of the features of $t$, i.e., $E\subseteq F$ where $F=\{w_1,\ldots,w_l\}$, that is sufficient to imply the prediction. We focus on deriving \textit{robust explanations}, i.e., on extracting a subset $E$ of the text features $F$ which ensure that the neural network prediction remains invariant for any perturbation of the other features $F\setminus E$. Thus, the features in a robust explanation are \textit{sufficient to imply the prediction} that we aim to explain, a clearly desirable feature for a local explanation. In particular, we focus on explanations that are \textit{robust w.r.t.\ bounded perturbations in the embedding space of the input text}. 
We extract word-level explanations by means of word embeddings: we note that OREs work, without further extensions, with diverse representations (e.g., sentence-level, characters-level, etc.). For a word $w \in W$, with embedding $x_w=\mathcal{E}(w)$ we denote with $\mathcal{B}(w)\subseteq \mathbb{R}^d$ a generic set of word-level perturbations. We consider the following kinds of perturbation sets, depicted also in Fig.~\ref{fig:knn-knn-box}. 

\vspace{.1cm}\noindent \textbf{$\epsilon$-ball}: $\mathcal{B}(w)=\{x \in \mathbb{R}^d \mid \|x - x_w\|_p \leq \epsilon \}$, for some $\epsilon>0$ and $p>0$. This is a standard measure of {local robustness} in computer vision, where $\epsilon$-variations are interpreted as  manipulations of the pixel intensity of an image. It has also been adopted in early NLP robustness works~\cite{miyato2016adversarial}, but then replaced with better representations based on actual word replacements and their embeddings, see below. 



\vspace{.1cm}\noindent \textbf{$k$-NN box closure}: $\mathcal{B}(w) = BB(\mathcal{E}(NN_k(w)))$, where $BB(X)$ is the minimum bounding box for set $X$; for a set $W'\subseteq W$, $\mathcal{E}(W') = \bigcup_{w' \in W'}\{\mathcal{E}(w')\}$; and $NN_k(w)$ is the set of the $k$ closest words to $w$ in the embedding space, i.e., words $w'$ with smallest $d(x_w, \mathcal{E}(w'))$, where $d$ is a valid notion of distance between embedded vectors, such as $p$-norms or cosine similarity\footnote{even though the box closure can be calculated for any set of embedded words.}. This provides an over-approximation of the $k$-NN convex closure, for which constraint propagation (and thus robustness checking) is more efficient~\cite{jia2019certified,huang2019achieving}. 

For some word-level perturbation $\mathcal{B}$, set of features $E\subseteq F$, and input text $t$ with embedding $(x_1,\ldots, x_l)$, we denote with $\mathcal{B}_E(t)$ the set of \textit{text-level} perturbations obtained from $t$ by keeping constant the features in $E$ and perturbing the others according to $\mathcal{B}$:
\begin{multline}\label{eq:text-pert}
    \mathcal{B}_E(t) = \{ (x'_1, \ldots, x'_l) \in \mathbb{R}^{l\cdot d} \mid 
    x'_w=x_w \text{ if } w \in E; \\ x'_{w} \in \mathcal{B}(w) \text{ otherwise} \}.
\end{multline}

A robust explanation $E\subseteq F$ ensures prediction invariance for any point in  $\mathcal{B}_E(t)$, i.e., any perturbation (within $\mathcal{B}$) of the features in $F\setminus E$.
\begin{definition}[Robust Explanation]\label{def:RE}
For a text $t=(w_1, \ldots, w_l)$ with embedding $x=\mathcal{E}(t)$, word-level perturbation $\mathcal{B}$, and classifier $M$, a subset $E\subseteq F$ of the features of $t$ is a \emph{robust explanation} iff
\begin{equation}\label{eq:robust}
    \forall x' \in \mathcal{B}_{E}(t) . \  M(x') = M(x). 
\end{equation}
We denote~\eqref{eq:robust} with predicate $\mathsf{Rob}_{M,x}(E)$.
\end{definition}

\noindent\textbf{Optimal Robust Explanations (OREs)} While robustness is a desirable property, it is not enough alone to produce useful explanations. Indeed, we can see that an explanation $E$ including all the features, i.e., $E=F$, trivially satisfies Definition~\ref{def:RE}. Typically, one seeks short explanations, because these can generalise to several instances beyond the input $x$ and are easier for human decision makers to interpret. We thus introduce \textit{optimal robust explanations (OREs)}, that is, explanations that are both robust and optimal w.r.t.\ an arbitrary cost function that assign a penalty to each word.

\begin{definition}[Optimal Robust Explanation]\label{def:ORE}
Given a cost function $\C: W \rightarrow \RR{^+}$, and for $t=(w_1, \ldots, w_l)$, $x$, $\mathcal{B}$, and $M$ as in Def.~\ref{def:RE}, a subset $E^*\subseteq F$ of the features of $t$ is an \emph{ORE} iff
\begin{equation}\label{eq:ore}
E^* \in \underset{E\subseteq F}{\argmin} \ \sum_{w \in E} \C(w) \text{ s.t. } \mathsf{Rob}_{M,x}(E).
\end{equation}
\end{definition}
Note that~\eqref{eq:ore} is always feasible, because its feasible set always includes at least the trivial explanation $E=F$. 
A special case of our OREs is when $\C$ is \textit{uniform} (it assigns the same cost to all words in $t$), in which case $E^*$ is (one of) the \textit{robust explanations of smallest size}, i.e., with the least number of words. 
\\

\noindent\textbf{Relation with Abductive Explanations} 
Our OREs have similarities with the \textit{abduction-based explanations (ABEs)} of~\cite{IgnatievNM19} in that they also derive minimal-cost explanations with robustness guarantees. For an input text $t=(w_1, \ldots, w_l)$, let $C = \bigwedge_{i=1}^l \chi_{i} = x_{w_i}$ be the \textit{cube} representing the embedding of $t$, where $\chi_{i}$ is a variable denoting the $i$-th feature of $x$. Let $\Net$ represent the logical encoding of the classifier $M$, and $\hat{y}$ be the formula representing the output of $\Net$ given $\chi_{1}, \ldots, \chi_l$. 
\begin{definition}[\cite{IgnatievNM19}]
An \emph{abduction-based explanation (ABE)} is a minimal cost subset $C^*$ of $C$ such that $C^* \wedge \Net \models \hat{y}$.
\end{definition}

Note that the above entailment is equivalently expressed as $C^* \models (\Net \rightarrow \hat{y})$. Let $B = \bigwedge_{i=1}^l \chi_{i} \in \mathcal{B}(w_i)$ be the constraints encoding our perturbation space. Then, the following proposition shows that OREs can be defined in a similar abductive fashion and also in terms of \textit{minimum satisfying assignments (MSAs)}~\cite{DilligDMA12}. 
In this way, we can derive OREs via analogous algorithms to those used for ABEs~\shortcite{IgnatievNM19} and MSAs~\cite{DilligDMA12}, 
as explained in Section~\ref{solutionAlgs}. Moreover, we find that every ORE can be formulated as a prime implicant~\cite{IgnatievNM19}, a property that connects our OREs with the notion of sufficient reason introduced in~\cite{darwiche2020reasons}. 
    
\begin{proposition}\label{prop:ore_abe}
Let $E^*$ be an ORE and $C^*$ its constraint encoding. Define $\phi \equiv (B \wedge \Net) \rightarrow \hat{y}$. Then, all the following definitions apply to $C^*$:
\begin{compactenum}
    \item $C^*$ is a minimal cost subset of $C$ such that $C^* \models \phi$.
    \item $C^*$ is a minimum satisfying assignment for $\phi$.
    \item $C^*$ is a prime implicant of $\phi$.
\end{compactenum}
\end{proposition}
\begin{proof}
See supplement
\end{proof}

The key difference with ABEs is that our OREs are robust to \textit{bounded} perturbations of the excluded features, while ABEs must be robust to \textit{any} possible perturbation. 
This is an important difference because it is hard (often impossible) to guarantee prediction invariance w.r.t.\ the entire input space when this space is continuous and high-dimensional, like in our NLP embeddings. In other words, if for our NLP tasks we allowed any word-level perturbation as in ABEs, in most cases the resulting OREs will be of the trivial kind, $E^*=F$ (or $C^*=C$), and thus of little use.  
For example, if we consider $\epsilon$-ball perturbations and the review \texttt{``the gorgeously elaborate continuation of the lord of the rings''}, the resulting smallest-size explanation is of the trivial kind (it contains the whole review) already at $\epsilon=0.1$. 
\\

\noindent\textbf{Exclude and include constraints} We further consider OREs $E^*$ derived under constraints that enforce specific features $F'$ to be included/excluded from the explanation: 
\begin{equation}\label{eq:core}
E^* \in \underset{E\subseteq F}{\argmin} \ \sum_{w \in E} \C(w) \text{ s.t. } \mathsf{Rob}_{M,x}(E) \wedge \phi(E),
\end{equation}
where $\phi(E)$ is one of $F' \cap E = \emptyset$ (\textit{exclude}) and $F' \subseteq E$ (\textit{include}). 
Note that adding \textit{include} constraints doesn't affect the feasibility of our problem\footnote{because the feasible region of~\eqref{eq:core} always contains at least the explanation $E^*\cup F'$, where $E^*$ is a solution of~\eqref{eq:ore} and $F'$ are the features to include. See Def.~\ref{def:RE}.}.
Conversely, \textit{exclude} constraints might make the problem infeasible when the features in $F'$ don't admit perturbations, i.e., they are necessary for the prediction, and thus cannot be excluded. Such constraints can be easily accommodated by any solution algorithm for non-constrained OREs: for \textit{include} ones, it is sufficient to restrict the feasible set of explanations to the supersets of $F'$; for \textit{exclude} constraints, we can manipulate the cost function so as to make any explanation with features in $F'$ strictly sub-optimal w.r.t.\ explanations without\footnote{That is, we use cost $\C'$ such that $\forall_{w \in F\setminus F'}\C'(w)=\C(w)$ and $\forall_{w' \in F'}\C'(w') > \sum_{w \in F\setminus F'} \C(w)$. The ORE obtained under cost $\C'$ might still include features from $F'$, which implies that~\eqref{eq:core} is infeasible (i.e., no robust explanation without elements of $F'$ exists).}. 

Constrained OREs enable two crucial use cases: \textit{detecting biased decisions}, and \textit{enhancing non-formal explainability frameworks}. 

\paragraph{Detecting bias} Following~\cite{darwiche2020reasons}, we deem a classifier decision \textit{biased} if it depends on protected features, i.e., a set of input words that should not affect the decision (e.g., a movie review affected by the director's name). In particular, a decision $M(x)$ is biased if we can find, {within a given set of text-level perturbations}, an input $x'$ that agrees with $x$ on all but protected features and such that $M(x)\neq M(x')$.
\begin{definition}\label{def:bias}
For classifier $M$, text $t$ with features $F$, protected features $F'$ and embedding $x=\mathcal{E}(t)$, decision $M(x)$ is \emph{biased} w.r.t.\ some word-level perturbation $\mathcal{B}$, if 
$$\exists x' \in \mathcal{B}_{F\setminus F'}(t) . M(x) \neq M(x').$$ 
\end{definition}
The proposition below allows us to use exclude constraints to detect bias. 
\begin{proposition}\label{prop:dec_bias}
For $M$, $t$, $F$, $F'$, $x$ and $\mathcal{B}$ as per Def.~\ref{def:bias}, \emph{decision $M(x)$ is biased iff~\eqref{eq:core} is infeasible under $F' \cap E = \emptyset$.} 
\end{proposition}
\begin{proof}
See supplement
\end{proof}


\paragraph{Enhancing non-formal explainers} The local explanations produced by heuristic approaches like LIME or Anchors do not enjoy the same robustness/invariance guarantees of our OREs. We can use our approach to \textit{minimally extend} (w.r.t.\ the chosen cost function) any non-robust local explanation $F'$ in order to make it robust, by solving~\eqref{eq:core} under the \textit{include} constraint $F' \subseteq E$. 
In particular, with a uniform $\C$, our approach would identify the smallest set of extra words that make $F'$ robust. Being minimal/smallest, such an extension retains to a large extent the original explainability properties. 

\paragraph{Relation with Anchors} Anchors~\cite{ribeiro2018anchors} are a state-of-the-art method for ML explanations. Given a perturbation distribution $\mathcal{D}$, classifier $M$ and input $x$, an anchor $A$ is a predicate over the input features such that $A(x)$ holds and $A$ has high \textit{precision} and \textit{coverage}, defined next.
\begin{equation} \label{eq:prec-cov}
    \mathsf{prec}(A) =  \Pr_{\mathcal{D}(x'\mid A(x'))}(M(x)=M(x')); \ 
    \mathsf{cov}(A) = \Pr_{\mathcal{D}(x')}(A(x'))
\end{equation}
In other words, $\mathsf{prec}(A)$ is the probability that the prediction is invariant for any perturbation $x'$ to which explanation $A$ applies. In this sense, precision can be intended as a robustness probability. $\mathsf{cov}(A)$ is the probability that explanation $A$ applies to a perturbation. To discuss the relation between Anchors and OREs, for an input text $t$, consider an arbitrary distribution $\mathcal{D}$ with support in $\mathcal{B}_{\emptyset}(t)$ (the set of all possible text-level perturbations),  see~\eqref{eq:text-pert}; and consider anchors $A$ defined as subsets $E$ of the input features $F$, i.e., $A_E(x)=\bigwedge_{w \in E} x_w = \mathcal{E}(w)$. Then, our OREs enjoy the following properties.
\begin{proposition}\label{prop:anchor1}
If $E$ is a robust explanation, then \mbox{$\mathsf{prec}(A_E)=1$}. 
\end{proposition}
\begin{proof}
See supplement
\end{proof}

Note that when $\mathcal{D}$ is continuous, $\mathsf{cov}(A_E)$ is always zero unless $E=\emptyset$\footnote{in which case $\mathsf{cov}(A_{\emptyset})=1$ (as $A_{\emptyset}=\mathsf{true}$). Indeed, for $E\neq\emptyset$, the set $\{x' \mid A_E(x')\}$ has $|E|$ fewer degrees of freedom than the support of $\mathcal{D}$, and thus has both measure and coverage equal to zero.}. We thus illustrate the next property assuming that $\mathcal{D}$ is discrete (when $\mathcal{D}$ is continuous, the following still applies to any empirical approximation of $\mathcal{D}$).

\begin{proposition}\label{prop:anchor2}
If $E\subseteq E'$, then $\mathsf{cov}(A_E)\geq \mathsf{cov}(A_{E'})$.
\end{proposition}
\begin{proof}
See supplement
\end{proof}
The above proposition suggests that using a uniform $\C$, i.e., minimizing the explanation's length, is a sensible strategy to obtain high-coverage OREs. 

\section{Solution Algorithms}
\label{solutionAlgs}
We present two solution algorithms to derive OREs, respectively based on the hitting-set (HS) paradigm of~\cite{IgnatievNM19} and the MSA algorithm of~\cite{DilligDMA12}. Albeit different, both algorithms rely on repeated entailment/robustness checks $B \wedge E \wedge \Net \models \hat{y}$ for a candidate explanation $E\subset C$. For this check, we employ two state-of-the-art neural network verification tools, Marabou~\cite{KatzHIJLLSTWZDK19} and Neurify \cite{wang2018efficient}: they both give provably correct answers and, when the entailment is not satisfied, produce a counter-example $x'\in \mathcal{B}_E(t)$, i.e., a perturbation that agrees with $E$ and such that $B \wedge C' \wedge \Net \not\models \hat{y}$, where $C'$ is the cube representing $x'$. 
We now briefly outline the two algorithms. A more detailed discussion (including the pseudo-code) is available in the supplement. 
\\

\noindent\textbf{Minimum Hitting Set}
For a counterexample $C'$, let $I'$ be the set of feature variables where $C'$ does not agree with $C$ (the cube representing the input). Then, every explanation $E$ that satisfies the entailment must hit all such sets $I'$ built for any counter-examples $C'$~\cite{IgnatievPM16}. Thus, the HS paradigm iteratively checks candidates $E$ built by selecting the subset of $C$ whose variables form a minimum HS (w.r.t.\ cost $\C$) of said $I'$s. 
However, we found that this method often struggles to converge for our NLP models, especially with large perturbations spaces (i.e., large $\epsilon$ or $k$). We solved this problem by extending the HS approach with a sub-routine that generates batches of \textit{sparse adversarial attacks} for the input $C$. 
This has a two-fold benefit: 1) we reduce the number of entailment queries required to produce counter-examples, and 2) sparsity results in small $I'$ sets, which further improves convergence. 

\noindent\textbf{Minimum Satisfying Assignment}
This algorithm exploits the duality between MSAs and maximum universal subsets (MUSs): for cost $\C$ and formula $\phi \equiv (B \wedge \Net) \rightarrow \hat{y}$, an MUS $X$ is a set of variables with maximum $\C$ such that $\forall X. \phi$, which implies that $C\setminus X$ is an MSA for $\phi$~\cite{DilligDMA12} and, in turn, an ORE. Thus, the algorithm of~\cite{DilligDMA12} focuses on deriving an MUS, and it does so in a recursive branch-and-bound manner, where each branch adds a feature to the candidate MUS. Such an algorithm is exponential in the worst-case, but we mitigated this by selecting a good ordering for feature exploration and performing entailment checks to rule out features that cannot be in the MUS (thus reducing the search tree). 

\begin{figure*}
\centering
\includegraphics[width=\linewidth]{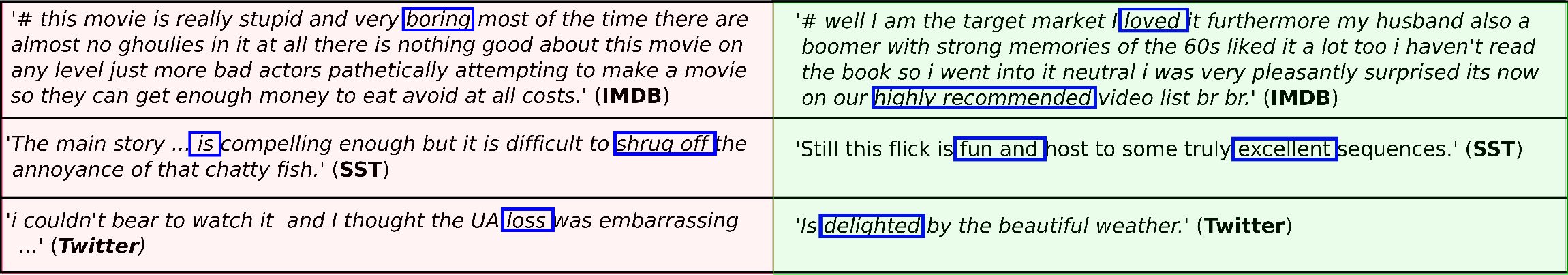}\caption{
OREs for IMDB, SST and Twitter datasets (all the texts are correctly classified). Models employed are FC with 50 input words each with accuracies respectively $0.89$, $0.77$ and $0.75$. OREs are highlighted in blue. Technique used is kNN boxes with k=$15$. 
}
\label{fig:good-OREs}
\end{figure*}

\section{Experimental Results}
\label{expres}

\noindent\textbf{Settings}
We have trained fully connected (FC) and convolutional neural networks (CNN) models on sentiment analysis datasets that differ in the input length and difficulty of the learning task\footnote{Experiments were parallelized on a server with two $24$-core Intel Xenon $6252$ processors and $256$GB of RAM, but each instance is single-threaded and can be executed on a low-end laptop.}.
We considered $3$ well-established benchmarks for sentiment analysis, namely SST~\cite{socherrecursive:2013}, Twitter~\cite{go2009twitter} and  IMDB~\cite{maas-EtAl:2011:ACL-HLT2011} datasets. From these, we have chosen $40$ representative input texts, balancing \textit{positive} and \textit{negative} examples. 
Embeddings are pre-trained on the same datasets used for classification~\cite{chollet2015keras}. 
Both the HS and MSA algorithms have been implemented in Python and use Marabou~\cite{KatzHIJLLSTWZDK19} and Neurify~\cite{wang2018efficient} to answer robustness queries\footnote{Marabou is fast at verifying ReLU FC networks, but it becomes memory intensive with CNNs. 
On the other hand, the symbolic interval analysis of Neurify is more efficient for CNNs. A downside of Neurify is that it is less flexible in the constraint definition (inputs have to be represented as squared bi-dimensional grids, thus posing problems for NLP inputs which are usually specified as 3-d tensors).}. 
In the experiments below, we opted for the kNN-box perturbation space, as we found that the $k$ parameter was easier to interpret and tune than the $\epsilon$ parameter for the $\epsilon$-ball space, and improved verification time. Further details on the experimental settings, including a selection of $\epsilon$-ball results, are given in the supplement.
\\

\begin{figure}
\centering
\includegraphics[width=\linewidth]{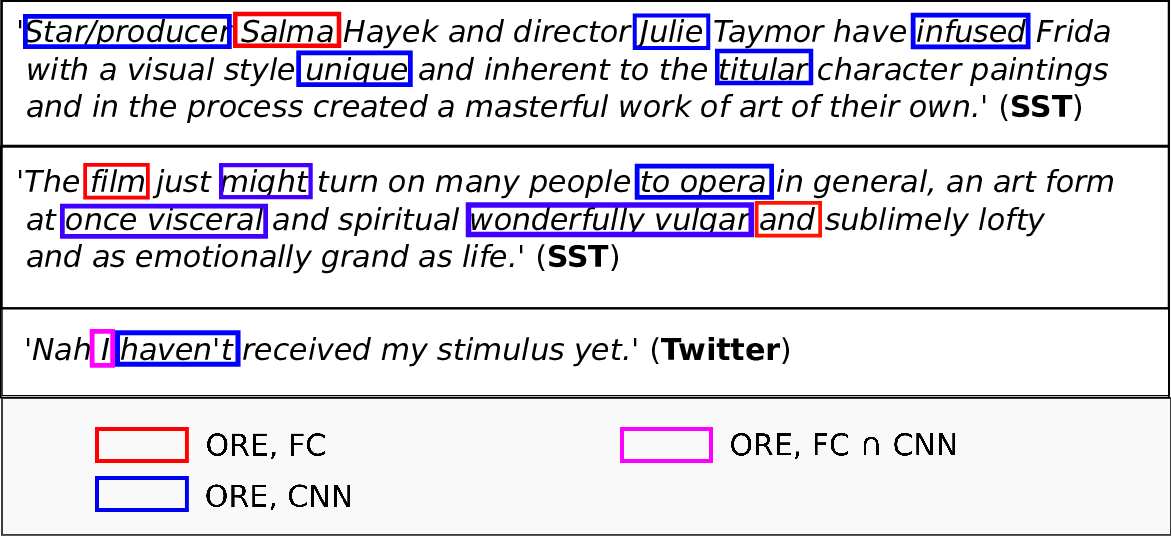}\caption{Comparison of OREs for SST and Twitter texts on FC (red) vs CNN (blue) models (common words in magenta). The first two are \textit{positive} reviews, the third is \textit{negative} (all correctly classified). Accuracies of FC and CNN models are, respectively, $0.88$ and $0.89$ on SST, $0.77$ on Twitter. Models have input length of $25$ words, OREs are extracted with kNN boxes (k=$25$).
}
\label{fig:fc-vs-cnn}
\end{figure}

\begin{figure}
\centering
\includegraphics[width=\linewidth]{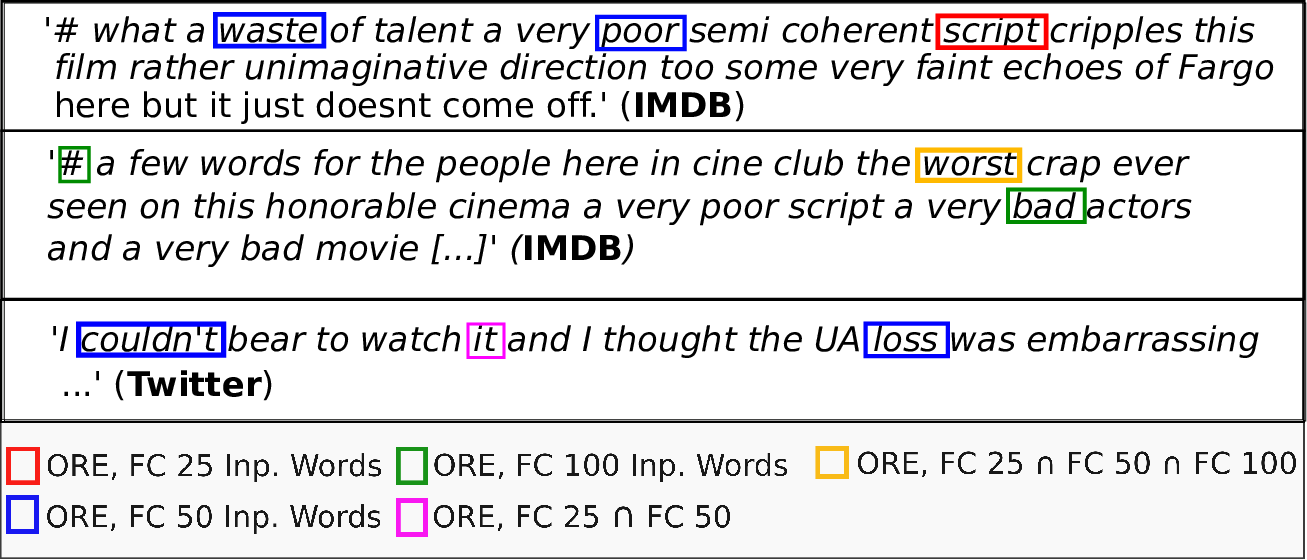}\caption{Comparison of OREs on \textit{negative} IMDB and Twitter inputs for FC models. The first and third examples are trained with 25 (red) VS 50 (blue) input words (words in common to both OREs are in magenta). The second example further uses an FC model trained with 100 input words (words in common to all three OREs are in orange). Accuracy is respectively $0.7$ and $0.77$ and $0.81$ for IMDB, and $0.77$ for both Twitter models. All the examples are classified correctly. OREs are extracted with kNN boxes (k=$25$). 
}
\label{fig:low-vs-high-accuracy}
\end{figure}

\noindent
\textbf{Effect of classifier's accuracy and robustness.}
We find that our approach generally results in meaningful and compact explanations for NLP. In Figure \ref{fig:good-OREs}, we show a few OREs extracted for \textit{negative} and \textit{positive} texts, where the returned OREs are both concise and semantically consistent with the predicted sentiment.
However, the quality of our OREs depends on that of the underlying classifier. Indeed, enhanced models with better accuracy and/or trained on longer inputs tend to produce higher quality OREs. We show this in Figures~\ref{fig:fc-vs-cnn} and \ref{fig:low-vs-high-accuracy}, where 
we observe that enhanced models tend to result in more semantically consistent explanations. For lower-quality models, some OREs include seemingly irrelevant terms (e.g., \textit{``film''}, \textit{``and''}), thus exhibiting 
shortcomings of the classifier. 
\\

\begin{figure}
    \centering
    \includegraphics[width=\linewidth]{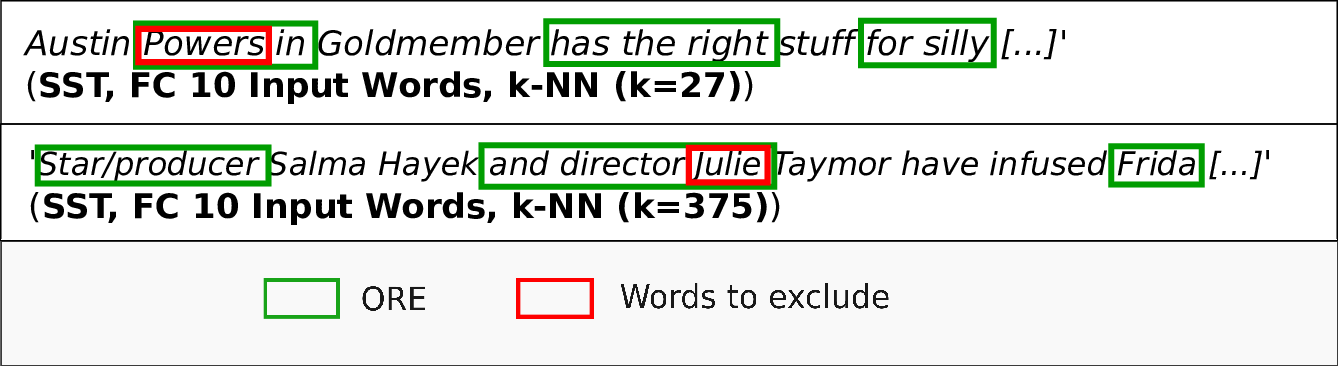}
    \caption{Two examples of decision bias from an FC model with an accuracy of $0.80$.}
    \label{fig:ores-decision-bias}
\end{figure}

\noindent\textbf{Detecting biases}
As per Prop.~\ref{prop:dec_bias}, we applied exclude constraints to detect biased decisions. In Figure \ref{fig:ores-decision-bias}, we provide a few example instances exhibiting such a bias, i.e., where \textit{any} robust explanation contains at least one protected feature. 
These OREs include proper names 
that shouldn't constitute a sufficient reason for the model's classification. When we try to exclude proper names, no robust explanation exists, indicating that a decision bias exists.
\\



\begin{figure}
\centering
\includegraphics[width=\linewidth]{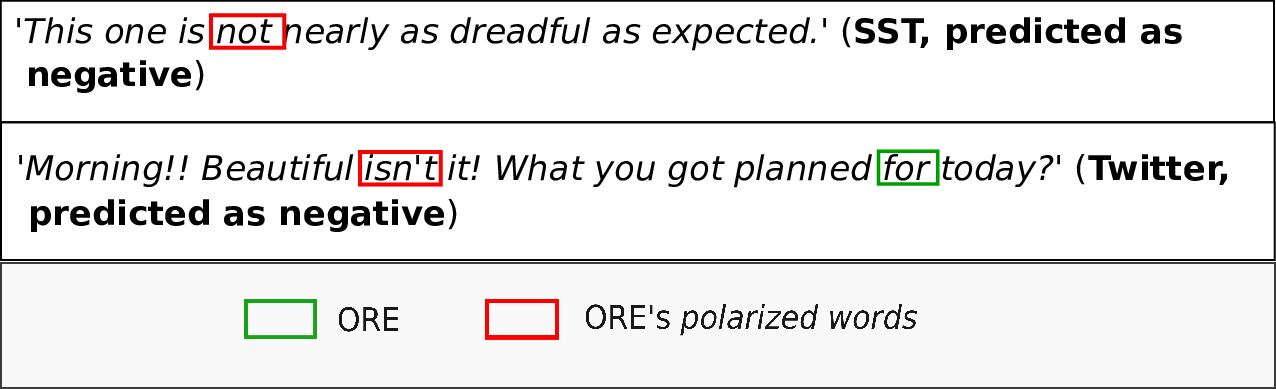}\caption{Two examples of over-sensitivity to polarized terms (in red). Other words in the OREs are highlighted in green. Models used are FC with 25 input words (accuracy $0.82$ and  $0.74$). Method used is kNN with k respectively equal to $8$ and $10$.}
\label{fig:over-sensitivity}
\end{figure}

\noindent\textbf{Debugging prediction errors} An important use-case for OREs is when a model commits
a \textit{misclassification}. Misclassifications in sentiment analysis tasks usually depend on over-sensitivity of the model to polarized terms. In this sense,  knowing a minimal, sufficient reason behind the model's prediction can be useful to debug it. As shown in the first example in Figure~\ref{fig:over-sensitivity}, the model cannot recognize the \textit{double negation} constituted by the terms \textit{not} and \textit{dreadful} as a syntax construct, hence it exploits the negation term \textit{not} to classify the review as \textit{negative}.
\\

\noindent\textbf{Comparison to Anchors}
We evaluate the robustness of Anchors for FC and CNN models on the SST and Twitter datasets\footnote{Accuracies are $0.89$ for FC+SST, $0.82$ for FC+Twitter, $0.89$ for CNN+SST, and $0.77$ for CNN+Twitter.}. 
To compute robustness, we assume a kNN-box perturbation space $\mathcal{B}$ with $k=15$ for FC and $k=25$ for CNN models. To extract Anchors, we set $\mathcal{D}$ to the standard perturbation distribution of~\cite{ribeiro2018anchors}, defined by a set of context-wise perturbations generated by a powerful language model. Thus defined $\mathcal{B}$s are small compared to the support of $\mathcal{D}$,
and so one would expect high-precision Anchors to be relatively robust w.r.t.\ said $\mathcal{B}$s. 
On the contrary, the Anchors extracted for the FC models attain an average precision of $0.996$ on SST and $0.975$ on Twitter, but only $12.5\%$ of them are robust for the SST case and $7.5\%$ for Twitter. With CNN models, high-quality Anchors are even more brittle: $0\%$ of Anchors are robust on SST reviews and $5.4\%$ on Twitter, despite an average precision of $0.995$ and $0.971$, respectively.

\begin{figure}
\centering
\includegraphics[height=3.8cm,width=\linewidth]{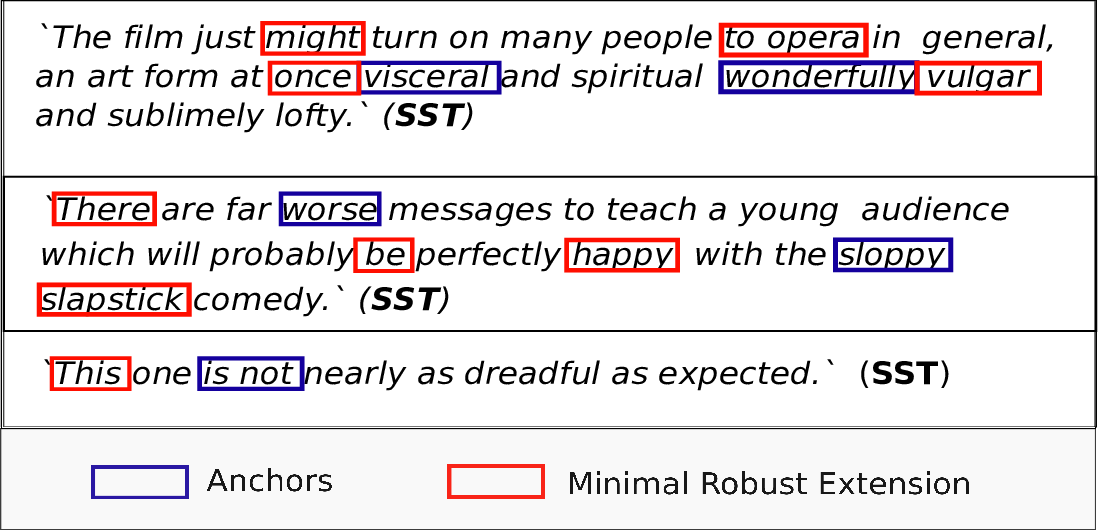}\caption{
Examples of Anchors explanations (in blue) along with the minimal extension required to make them robust (in red).
Examples are 
classified (without errors)
with a 25-input-word CNN (accuracy $0.89$). 
OREs are extracted for kNN boxes and k=$25$.}
\label{fig:anchors-completion}
\end{figure}

We remark, however, that Anchors are not designed to provide such robustness guarantees. 
Our approach becomes useful in this context, because it can \textit{minimally extend} any local explanation to make it robust, by using \textit{include constraints} as explained in Section 3. In Figure~\ref{fig:anchors-completion} we show a few examples of how, starting from non-robust Anchors explanations, our algorithm can find the minimum number of words to make them provably robust.

\section{Conclusions}

We have introduced optimal robust explanations (OREs) and applied them to enhance interpretability of NLP models. OREs provide concise and sufficient reasons for a particular prediction, as they are guaranteed to be both minimal w.r.t.\ a given cost function and robust, in that the prediction is invariant for any bounded replacement of the left-out features. We have presented two solution algorithms that build on the relation between our OREs, abduction-based explanations and minimum satisfying assignments. We have demonstrated the usefulness of our approach on widely-adopted sentiment analysis tasks, providing explanations for neural network models beyond reach for existing formal explainers. 
Detecting biased decisions, debugging misclassifications, and repairing non-robust explanations are some of key use cases  that our OREs enable. Future research plans include exploring more general classes of perturbations beyond the embedding space.

\noindent\textbf{Acknowledgements}
This project has received funding from the European Research Council (ERC)
under the European Union’s Horizon 2020 research and innovation programme
(FUN2MODEL, grant agreement No.~834115) 
and the EPSRC Programme Grant on Mobile Autonomy (EP/M019918/1).

\bibliographystyle{named}
\bibliography{ijcai21}

\clearpage
\section{Appendix}
We structure the Appendix in the following way. We first provide proofs of the propositions in Section 3. Second, we give details (through the pseudo-code) of the Algorithms and sub-routines that were used to find Optimal Robust Explanations: in particular we describe the \textit{shrink} (used to improve MSA) and the Adversarial Attacks procedures (used to improve HS).
We then provide details on the datasets and the architectures that we have used in the Experimental Evaluation, and finally we report many examples of interesting OREs that we were able to extract with our methods, alongside with tables that complete the comparison between MSA and HS as described in the Experimental Evaluation Section.

\subsection{Proofs}
\paragraph{Proof of  Prop.~\ref{prop:dec_bias}}

Call $A=$ \textit{``$M(x)$ is biased''} and $B=$ \textit{``\eqref{eq:core} is infeasible under $F' \cap E = \emptyset$''}. 
Let us prove first that $B\rightarrow A$. Note that $B$ can be equivalently expressed as
$$\forall E \subseteq F . ( E \cap F' \neq \emptyset \vee \exists x' \in \mathcal{B}_E(t) . M(x)\neq M(x') )$$
If the above holds for all $E$ then it holds also for $E=F\setminus F'$, and so it must be that $\exists x' \in \mathcal{B}_{F\setminus F'}(t) . M(x)\neq M(x')$ because the first disjunct is clearly false for $E=F\setminus F'$.

We now prove $A\rightarrow B$ by showing that $\neg B \rightarrow \neg A$. Note that $\neg B$ can be expressed as
\begin{equation}\label{eq:biasproof1}
    \exists E \subseteq F . (E \cap F' = \emptyset \wedge \forall x' \in \mathcal{B}_E(t) . M(x) = M(x')),
\end{equation}
and $\neg A$ can be expressed as
\begin{equation}\label{eq:biasproof2}
    \forall x' \in \mathcal{B}_{F\setminus F'}(t) . M(x) = M(x').
\end{equation}
To see that \eqref{eq:biasproof1} implies \eqref{eq:biasproof2}, note that 
any $E$ that satisfies \eqref{eq:biasproof1} must be such that $E \cap F' = \emptyset$, which implies that $E \subseteq F\setminus F'$, which in turn implies that $\mathcal{B}_{F\setminus F'}(t) \subseteq \mathcal{B}_E(t)$. By \eqref{eq:biasproof1}, the prediction is invariant for any $x'$ in $\mathcal{B}_E(t)$, and so is for any $x'$ in $\mathcal{B}_{F\setminus F'}(t)$.

\paragraph{Proof of Prop.~\ref{prop:anchor1}} 
A robust explanation $E\subseteq F$ guarantees prediction invariance for any $x' \in \mathcal{B}_{E}(t)$, i.e., for any $x'$ (in the support of $\mathcal{D}$) to which anchor $A_E$ applies. 

\paragraph{Proof of Prop.~\ref{prop:anchor2}} 

For discrete $\mathcal{D}$ with pmf $f_{\mathcal{D}}$, we can express $\mathsf{cov}(A_E)$ as 
\begin{multline*}
    \mathsf{cov}(A_E) = \sum_{x' \in supp(\mathcal{D})} f_{\mathcal{D}}(x')\cdot \mathbf{1}_{A_E(x')} =\\
    \sum_{x' \in supp(\mathcal{D})} f_{\mathcal{D}}(x')\cdot 
    \prod_{w \in E} \mathbf{1}_{x'_w = \mathcal{E}(w)}
\end{multline*}
To see that, for $E'\supseteq E$, $\mathsf{cov}(A_{E'})\leq \mathsf{cov}(A_E)$, observe that $\mathsf{cov}(A_{E'})$ can be expressed as
\begin{multline*}
    \mathsf{cov}(A_{E'}) = \sum_{x' \in supp(\mathcal{D})} f_{\mathcal{D}}(x')\cdot 
    \prod_{w \in E'} \mathbf{1}_{x'_w = \mathcal{E}(w)} =\\
    \sum_{x' \in supp(\mathcal{D})} f_{\mathcal{D}}(x')\cdot 
    \prod_{w \in E} \mathbf{1}_{x'_w = \mathcal{E}(w)} \cdot \prod_{w \in E\setminus E'} \mathbf{1}_{x'_w = \mathcal{E}(w)}
\end{multline*}
and that for any $x'$, $\prod_{w \in E\setminus E'} \mathbf{1}_{x'_w = \mathcal{E}(w)} \leq 1$.

\paragraph{Proof of Prop.~\ref{prop:ore_abe}}
With abuse of notation, in the following we use $C^*$ to denote both an ORE and its logical encoding. 
\begin{enumerate}
    \item if $C^*$ is an ORE, then $\phi \equiv (B \wedge \Net) \rightarrow \hat{y}$ is true for \textit{any} assignment $x'$ of the features not in $C^*$. In particular, $\phi$ is trivially satisfied for any $x'$ outside the perturbation space $B$, and, by Definition~\ref{def:RE}, is satisfied for any $x'$ within the perturbation space.
    \item As also explained in~\cite{DilligDMA12}, finding an optimal $C^*$ such that $C^* \models \phi$ is equivalent to finding an MSA $C^*$ for $\phi$. We should note that $C^*$ is a special case of an MSA, because the possible assignments for the variables in $C^*$ are restricted to the subsets of the cube $C$.
    \item $C^*$ is said a prime implicant of $\phi$ if $C^* \models \phi$ and there are no proper subsets $C'\subset C^*$ such that $C' \models \phi$. This holds regardless of the choice of the cost $\C$, as long as it is additive and assigns a positive cost to each feature as per Definition~\ref{def:ORE}. Indeed, for such a cost function, any proper subset $C'\subset C^*$ would have cost strictly below that of $C^*$,  meaning that $C' \not\models \phi$ (i.e., is not robust) because otherwise, $C'$ (and not $C^*$) would have been (one of) the robust explanations with minimal cost. 
\end{enumerate}

\subsection{Optimal Cost Algorithms and Sub-Routines} \label{sec:hs-msa-full}

In this Section we provide a full description and the pseudo-code of the algorithms that for reason of space we were not able to insert in the main paper. We report a line-by-line  description of the HS procedure (Algorithm \ref{alg:hs}): we further describe how the adversarial attacks procedure is used to generate candidates that help the HS approach converge on hard instances, as reported in Section $4$. We then describe the algorithm to compute Smallest Cost Explanations (Algorithm \ref{MSA}). In Algorithm \ref{shrink}, we finally detail the \textit{shrink} procedure as sketched in Section 3. 

\noindent\textbf{Minimal Hitting-Sets and Explanations}
One way to compute optimal explanations against a cost function $C$, is through the hitting set paradigm~\cite{IgnatievNM19}, that exploits the relationship between diagnoses and conflicts~\cite{reiter1987theory}: the idea is to collect perturbations and to calculate on their indices a minimum hitting set (MHS) i.e., a minimum-cost explanation whose features are in common with all the others. We extend this framework to find a word-level explanation for non-trivial NLP models. 
At each iteration of Algorithm \ref{alg:hs}, a minimum hitting set $E$ is extracted (line 3) from the (initially empty, line 1) set $\Gamma$. If function \textit{Entails} evaluates to \textit{False} (i.e., the neural network $\Net$ is provably safe against perturbations on the set of features identified by $F \setminus F$) the procedure terminates and $E$ is returned as an ORE. Otherwise, (at least) one feasible attack is computed on $F \setminus E$ and added to $\Gamma$ (lines 7-8): the routine then re-starts. Differently from~\cite{IgnatievNM19}, as we have experienced that many OREs whose a large perturbation space - i.e. when $\epsilon$ or $k$ are large - do not terminate in a reasonable amount of time, we have extended the \textit{vanilla} hitting set approach by introducing \textit{SparseAttacks} function (line 7). At each iteration \textit{SparseAttacks} introduces in the hitting set $\Gamma$ a large number of sparse adversarial attacks on the set of features $F \setminus E$: it is in fact known~\cite{IgnatievPM16} that attacks that use as few features as possible help convergence on instances that are hard (intuitively, a small set is harder to ``hit'' hence contributes substantially to the optimal solution compared to a longer one)
\textit{SparseAttacks} procedure is based on random search and it is inspired by recent works in image recognition and malaware detection~\cite{croce2020sparse}: pseudo-code is reported in \ref{alg:adv}, while a detailed description follows in the next paragraph. 

\paragraph{Sparse Adversarial Attacks}
In Algorithm \ref{alg:adv} we present a method to generate sparse adversarial attacks against features (i.e., words) of a generic input text. $GeneratePerturbations(k, n, Q)$ (line 2) returns a random population of $n$ perturbations that succeed at changing $\Net$'s classification: for each successful attack $p$, a subset of $k$ out of $d$ features has been perturbed through a Fast Gradient Sign attack \footnote{\url{https://www.tensorflow.org/tutorials/generative/adversarial_fgsm}} (FGSM), while it is ensured that the point lies inside a convex region $Q$ which in our case will be the $\epsilon$ hyper-cube around the embedded text. If no perturbation is found in this way (i.e., population size of the atacks is zero, as in line 3), budget is decreased (line 4) and another trial of $GeneratePerturbations(k, n, Q)$ is performed (e.g., with few features as targets and a different random seed to guide the attacks). Function $AccuracyDrop(\Net,P)$ returns the best perturbation $a$ where $k$ is increasingly minimised (line 7). Algorithm terminates when either no attacks are possible (all the combinations of features have been explored) or after fixed number of iterations has been performed (line 1).

\begin{algorithm}[]
\SetAlgoLined
\KwData{a network $\Net$,
the input text $t$, the initial set of features F, a~network prediction $\E$ , a cost function $\C$ against which the explanation is minimised}
\KwResult{an optimal ORE E}
\SetAlgoVlined
$\Gamma = \emptyset$ \\
\While{true}{
    $E = MinimumHS(\Gamma, \C)$ \\
    
    \eIf{$Entails(E, (\Net\wedge \mathcal{B}_{F \setminus E}(t)) \to \E)$}{
    \Return $E$  \\
   }{
    $A = SparseAttacks(E, \Net)$ \\
    $\Gamma = \Gamma \cup \{A\}$ \\
   }
 }
\caption{ORE computation via implicit hitting sets and sparse attacks
\label{alg:hs}}
\end{algorithm}

\begin{algorithm}[]
\SetAlgoLined
\KwData{$\Net$ - neural network model, $F$ - input text from feature space; $k \in \mathbb{N}^+_{\setminus \{0\}}$ - number of perturbations initially tested; $Q \subseteq F$ - (sub)set of features where perturbations are found; $n \in \mathbb{N}^+_{\setminus \{0\}}$ - number of elements generated at each iteration of the algorithm; budget - number of iterations allowed before stopping.} 
\While{$k>0 \ \wedge \ budget>0$}{
    $P \xleftarrow{} GeneratePerturbations(k, n, Q)$ \\
    \If{length(P) $==0$}{
        $budget \xleftarrow{} budget - 1$ \\
        continue
    }
    $a \xleftarrow{} \arg \max_{p \in P} AccuracyDrop(M, P)$ \\
    $k \xleftarrow{} k-1$, $budget \xleftarrow{} budget - 1$
    }
return $a$
 \caption{Computing a perturbation that is successful and minimises the number of features that are perturbed.}
\label{alg:adv}
\end{algorithm}

\noindent\textbf{Minimum Satisfying Assignment Explanations}
\label{MSAE}
This approach, based on the method presented in~\cite{DilligDMA12}, finds an explanation in the form of an MSA, for which in turn a maximum universal subset (MUS) is required. For a given cost function $\C$ and text $t$, an MUS is a universal subset $t'$ of words that maximises $\C(t')$. An MSA of the network $M$ w.r.t the text is precisely a satisfying assignment of the formula $\forall_{w \in t'} . M\rightarrow \E$ for some MUS $t'$. In other words, an MSA is $t\setminus t'$. 
The inputs to the MSA algorithm 
are: $\Net$ which represents the network $M$ in constraint form;
text $t$;
cost function $\C$ and prediction $\E$ for the input $t$. 
The algorithm first uses the reversed sort function for the text $t$ to optimize the search tree. The text is sorted by the cost of each word.
then uses the recursive MUS algorithm to compute an MUS $t'$. Finally, the optimal explanation ($t\setminus t'$) is returned.
 
\begin{algorithm}[ht]
\SetAlgoLined
\DontPrintSemicolon
\KwData{a list of bounded words $bW$, 
a network $\Net$, 
a set of candidate words $cW$, 
the input text $t$,
a cost function $\C$ against which the ORE is minimised,
a lower bound for MUS $L$,
a prediction $\E$ for the input}
\KwResult{a Maximum Universal Subset with respect to input text $t$}
\lIf{ $c W= \emptyset$ or $\C(cW) \leqslant L$}{ 
        \Return $\emptyset$
}
\SetAlgoVlined
$best = \emptyset$\;
choose $w \in cW$\; 
$bW = bW \cup \{w\}$, $constW = cW \setminus \{w\}$\; 
\If{$Entails(constW, (\Net\wedge \mathcal{B}_{F \setminus E}(constW)) \to \E)$}{
    $Y = mus(bW, \Net, shrink(\Net, bW, cW\setminus \{w\}), t, \C, L-\C(w), \E)$\;
    $cost = \C(Y) + \C(w)$\;
    \If {$cost > L$} {
        $best = Y \cup \{w\}$\;
        $L = cost$\;
    }
}
$Y = mus(bW \setminus \{w\}, \Net, cW\setminus \{w\}, t, \C, L,\E)$\;

\lIf {$\C(Y) > L$} {
    $best = Y$
}
\Return $best$
 \caption{MUS computation,\\ $mus(bW, \Net, cW, t, \C, L, \E)$}
 \label{mus}
\end{algorithm}

The inputs of the $mus$ algorithm are: a set of candidate words $cW$ that an MUS should be calculated for (equal to $t$ in the first recursive call), a set of bounded words $bW$ that may be part of an MUS, where $\forall_{w \in bW}$, $w$ may be limited by $\epsilon$-ball or $k$-NN box clousure, 
a lower bound $L$, the network $\Net$, a cost function $\C$, 
and a network prediction $\E$.
It returns a maximum-cost universal set for the network $\Net$ with respect to $t$, which is a subset of $cW$ with a cost greater than $L$, or the empty set when no such subset exists. The lower bound allows us to cut off the search when the current best result cannot be improved.
During each recursive call, if the lower bound cannot be improved, the empty set is returned (line 1). Otherwise, a word $w$ is chosen from the set of candidate words $cW$ and it is determined whether the cost of the universal subset containing word $w$ is higher than the cost of the universal subset without it (lines 5-12). Before definitively adding word $w$ to $bW$, we test whether the result is still satisfiable with $Entails$ (line 5) i.e. still an explanation. 
The $shrink$ method helps to reduce the set of candidate words by iterating through current candidates and checking using $Entails$ whether they are necessary. This speeds-up the algorithm (as there are fewer overall calls to $Entails$). The recursive call at line 6 computes the maximum universal subset of $\forall_{w \in bW} \Net \to \E$, with adjusted $cW$ and $L$ as necessary. Finally within this $if$ block, we compute the cost of the universal subset involving word $w$, and if it is higher than the previous bound $L$, we set the new lower bound to cost (lines 7-11). Lines 11-12 considers the cost of the universal subset \textit{not} containing word $w$, in case it has higher cost, and if so, updates $best$. 
Once one optimal explanation has been found, it is possible to compute \textit{all combinations} of the input that match that cost, and then use $Entails$ on each to keep only those that are also explanations.

\noindent\textbf{Comparing MHS and MSA}
The MSA-based approach uses MUS algorithm to find maximum universal subset and then finds a MSA for that MUS. MUS is a recursive branch-and-bound algorithm~\cite{DilligDMA12} that explores a binary tree structure. The tree consists of all the word appearing in the input cube. The MUS algorithm possibly explores an exponential number of universal subsets, however, the recursion can be cut by using right words ordering (i.e. words for which robustness query will answer false, consider words with the highest cost first) or with shrink method. MUS starts to work with a full set of candidate words, whereas the HS approach starts with an empty set of fixed words and tries to find an attack for a full set of bounded words. In each iteration step, the HS approach increases the set of fixed words and tries to find an attack. It is because a subset $t' \subseteq t$ is an MSA for a classifier $M$ with respect to input text $t$ iff $t'$ is a minimal hitting set of minimum falsifying set (see~\cite{IgnatievPM16} for details). To speed up the MSA algorithm, we use $shrink$ procedure which reduces the set of candidate words, and for non-uniform cost function, words ordering (words with the highest cost are considered as the first candidates), while HS-based approach uses $SparseAttacks$ routine to increase the hitting set faster.

\begin{algorithm}[ht]
\SetAlgoLined
\DontPrintSemicolon
\KwData{a network $\Net$, 
an input text $t$, 
a cost function $\C$ for the input $C$,
a prediction $\E$,}
\KwResult{A smallest cost explanation for network $\Net$ w.r.t. input text $t$}
$bW =\emptyset, cW = C, sce = \emptyset$\;
$textSortedByCost = sort(t)$\;
$maxus = mus(bW, \Net, cW, textSortedByCost, \C, 0, \E)$ \;
\ForEach{$c \in t$}{
    \If {$c \notin maxus$} {
        $sce = sce \cup c$
    }
}
\Return $sce$
\caption{Computing smallest cost explanation}
\label{MSA}
\end{algorithm}

\begin{algorithm}[ht]
\SetAlgoLined
\DontPrintSemicolon
\KwData{a list of bounded words $bW$, 
a network $\Net$, 
a set of candidate words $cW$, 
a text $t$,
a cost function $\C$,
a lower bound $L$,
a prediction $\E$ for the input}
\KwResult{A set of the essential candidate words eW}
$eW = cW$\;
\ForEach{$word \in cW$}{
    $eW = eW\setminus\{word\}$\;
    $bW = bW \cup \{word\}$\;
    $constW = C \setminus bw$\;
    \If{$Entails(constW, (\Net\wedge \mathcal{B}_{F \setminus E}(cW)) \to \E)$}{
    $eW = eW \cup \{word\}$\;
    }
    $bW = bW\setminus\{word\}$\;
}
\Return $eW$
\caption{$shrink$ algorithm\\ $shrink(bW, \Net, cW, C, \C, L, \E)$}
\label{shrink}
\end{algorithm}

\noindent\textbf{Excluding words from MSA} To exclude specific words from a smallest explanation we add one extra argument to the MSA algorithm input: the $bW$ which represents bounded words. In this case the set $cW = t \setminus bW$. From now on the procedure is the standard one.

\subsection{Details on the Experimental Results} \label{sec:details-exp}
\subsubsection{Datasets and Test Bed}
As mentioned in the Experimental Evaluation Section, we have tested MSA and HS approaches for finding optimal cost explanations respectively on the SST, Twitter and IMDB datasets. For each task, we have selected a sample of $40$ input texts that maintain classes balanced (i.e., half of the examples are \textit{negative}, half are \textit{positive}). Moreover, we  inserted inputs whose polarity was exacerbated (either very \textit{negative} or very \textit{positive}) as well as more challenging examples that machines usually misclassifies, like \textit{double negations} or mixed sentiments etc. Further details in Table \ref{tab:data}.

\subsubsection{Models Setup}
We performed our experiments on FC and CNNs with up to $6$ layers and $20K$ parameters. FC are constituted by a stack of \textit{Dense} layers, while CNNs additionally employ \textit{Convolutional} and \textit{MaxPool} layers: for both CNNs and FC the decision is taken through a \textit{softmax} layer, with \textit{Dropout} that is addedd after each layer to improve generalization during the training phase. As regards the embeddings that the models equip, we experienced that the best trade-off between the accuracy of the network and the formal guarantees that we need to provide is reached with low-dimensional embeddings, thus we employed optimized vectors of dimension $5$ for each word in the embedding space: this is in line with the experimental evaluations conducted in \cite{patel2017towards}, where for low-order tasks such as sentiment analysis, compact embedding vectors allow to obtain good performances, as shown in Table \ref{tab:data}. We note that techniques such as retro-fitting \cite{faruqui2014retrofitting} could allow using more complex representations and might help with high-order tasks such as multi-class classification, where the quality of the embedding plays a crucial role in terms of accuracy. We will consider this as a future extension of the work. We report in Table \ref{tab:data} an overview of the models we used.

\begin{table*}[t]
\centering
Table \ref{tab:data}.1: Training
\\
\scalebox{1.0}{
\begin{tabular}{|c|c|c|c|}
\hline
 \textbf{} &  \small{\textbf{TWITTER}} & \small{\textbf{SST}} & \small{\textbf{IMDB}} \\ \hline
  \multirow{1}{*}{\small{\textbf{Inputs (Train, Test)}}} & \small{$ 1.55M, 50K$} &  \small{$117.22K,1.82K$} & \small{$25K, 25K$} \\ \cline{1-4}
  \multirow{1}{*}{\small{\textbf{Output Classes}}} & \small{$2$} &  \small{$2$} &  \small{$2$} \\ \cline{1-4}
  \multirow{1}{*}{\small{\textbf{Input Length (max, max. used)}}} & \small{$88$, $50$} &  \small{$52$, $50$} &  \small{$2315$, $100$} \\ \cline{1-4}  
   \multirow{1}{*}{\small{\textbf{Neural Network Models}}} & \small{FC, CNN} & \small{FC, CNN} & \small{FC, CNN} \\ \cline{1-4}
   \multirow{1}{*}{\small{\textbf{Neural Network Layers (min,max)}}} & \small{3,6} & \small{3,6} & \small{3,6} \\ \cline{1-4}
  \multirow{1}{*}{\small{\textbf{Accuracy on Test Set (min, max)}}} & \small{0.77, 0.81} & \small{0.82, 0.89} & \small{0.69, 0.81} \\ \cline{1-4}
  \multirow{1}{*}{\small{\textbf{Number of Networks Parameters (min,max)}}} & \small{$3K,18K$} & \small{$1.3K,10K$} & \small{$5K,17K$} \\ \cline{1-4}
\end{tabular} 
}
\newline
\vspace*{0.1 cm}
\newline
Table \ref{tab:data}.2: Explanations
\\
\scalebox{1.0}{
\begin{tabular}{|c|c|c|c|}
\hline
 \textbf{} &  \small{\textbf{TWITTER}} & \small{\textbf{SST}} & \small{\textbf{IMDB}} \\ \hline
   \multirow{1}{*}{\small{\textbf{Sample Size}}} & \small{$40$} &  \small{$40$} &  \small{$40$} \\ \cline{1-4}
   \multirow{1}{*}{\small{\textbf{Review Length (min-max)}}} & \small{10, 50} &  \small{10, 50} &  \small{25, 100} \\ \cline{1-4}
\end{tabular} 
}
\caption{
Datasets used for training/testing and extracting explanations. We report various metrics concerning the networks and the training phase (included accuracy on Test set), while in Table \ref{tab:data}.2 we report the number of texts for which we have extracted explanations along with the number of words considered when calculating OREs: samples were chosen to reflect the variety of the original datasets, i.e., a mix of long/short inputs equally divided into positive and negative instances.}
\label{tab:data}
\end{table*}

\begin{table}[]
\centering
\begin{small}
\begin{tabular}{|c|c|c|c|}
\hline
\textbf{\small{$\epsilon$}} & \begin{tabular}[c]{@{}c@{}}\textbf{\small{Explanation}}\\ \textbf{\small{Length}}\end{tabular} & \begin{tabular}[c]{@{}c@{}}\textbf{\small{MSA}}  \\ \textbf{\small{Execution Time}}\end{tabular} & \begin{tabular}[c]{@{}c@{}}\textbf{\small{HS}} \\ \textbf{\small{Execution Time}}\end{tabular} \\ \hline
\textbf{\small{0.01}} & $5\pm 5$ & $ 8.08\pm 7.9$ & $63.70\pm 63.69$ \\ \hline
\textbf{\small{0.05}} & $5.5\pm 4.5$ & $176.22\pm 175.92$ & $339.96\pm 334.66$ \\ \hline
\textbf{\small{0.1}} & $7.5\pm 2.5$ & $ 2539.75\pm 2539.14 $ & $3563.4 \pm 3535.84$ \\ \hline
\end{tabular}
\caption{Comparison between MSA and HS in terms of execution time for different values of $\epsilon$, and the corresponding explanation length.
}
\label{tab:mus-vs-hs-1-100}
\end{small}
\end{table}


\subsection{Additional Results} \label{sec:eps-results}
In this Section we provide a few interesting results that we couldn't add to the main paper.

\subsubsection{Additional kNN Results}
\begin{figure}
\centering
\includegraphics[width=\linewidth]{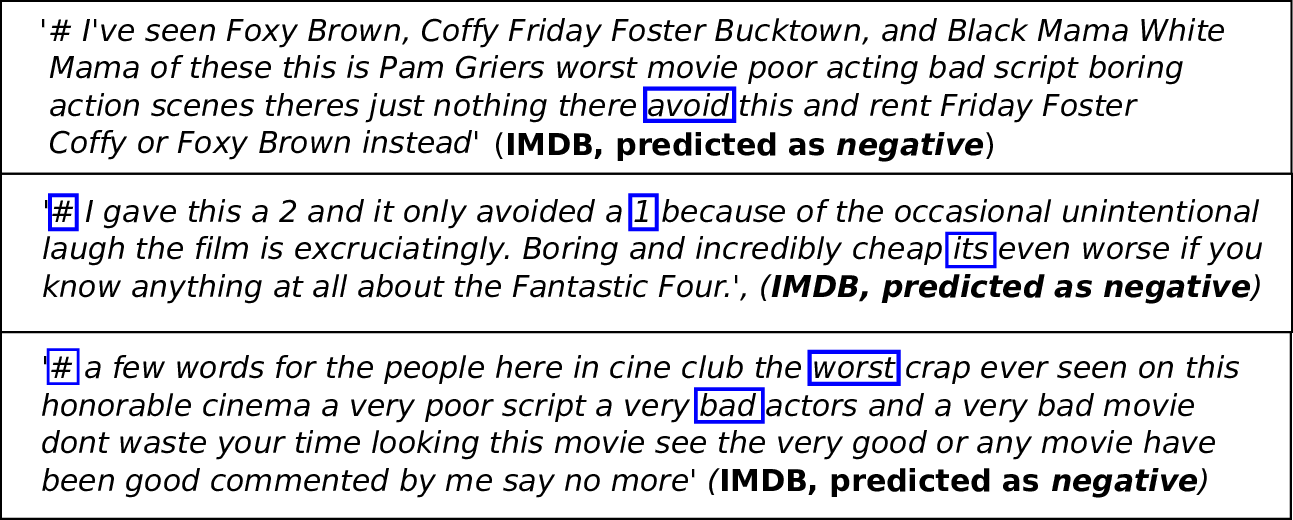}\caption{Examples of Optimal Robust Explanations - highlighted in blue -. OREs were extracted using kNN boxes with $25$ neighbors per-word: fixing words in an ORE guarantees the model to be locally robust. The examples come from the IMDB dataset, model employed is a FC network with $100$ input words (accuracy $0.81$).}
\label{fig:100inputs-IMDB-OREs}
\end{figure}

\begin{figure}
\centering
\includegraphics[width=\linewidth]{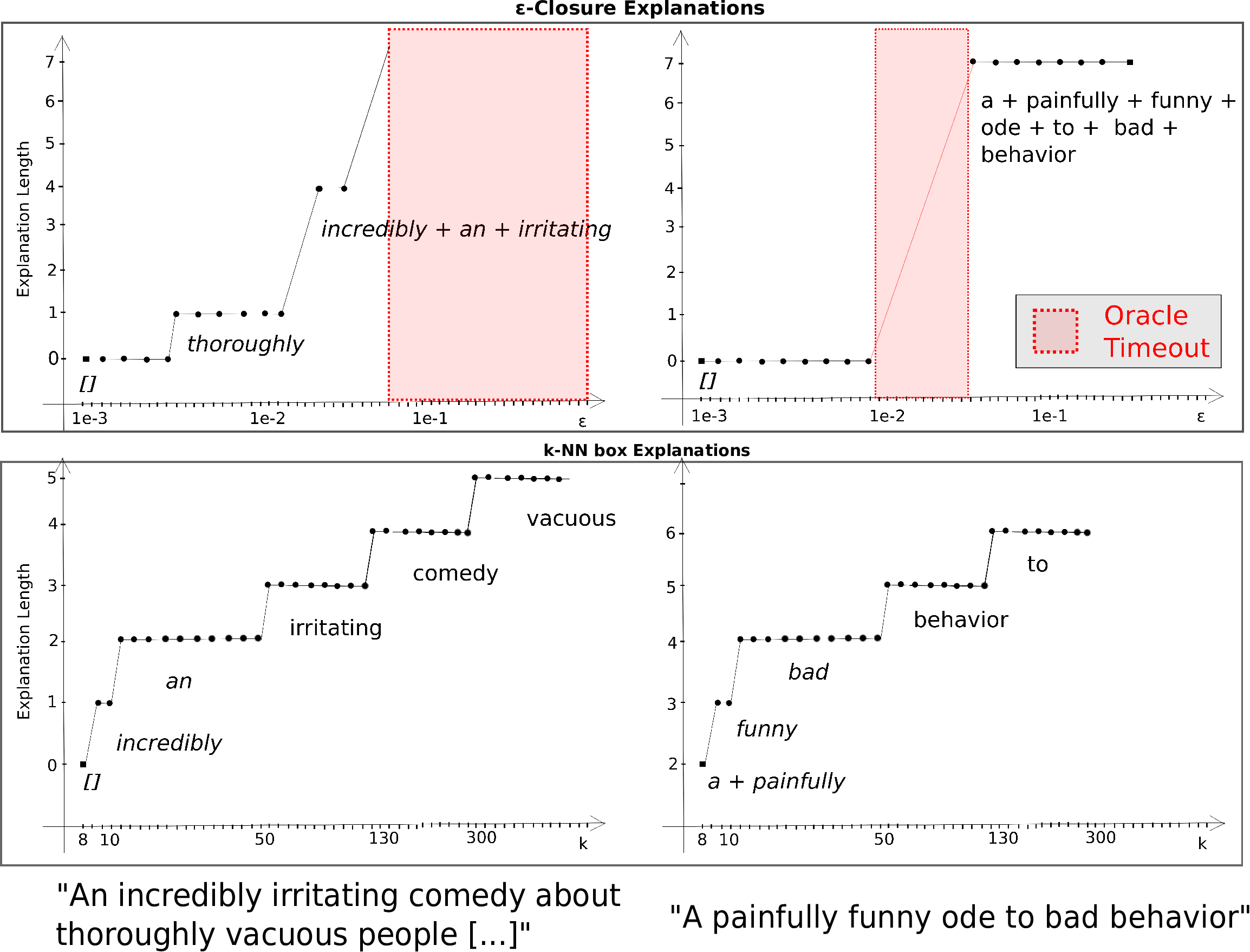}\caption{
How an explanation grows when either $\epsilon$ (top) or k (bottom) is increased. Model considered is a fully connected with 50 input words on SST dataset (0.89 accuracy). On the left a \textit{negative} review that is correctly classified, on the right a \textit{positive} review that is misclassified (i.e., the model's prediction is \textit{negative}). For specific ranges of $\epsilon$ the Oracle cannot extract an explanation (timeout, highlighted in red).}
\label{fig:hard-instances}
\end{figure}

As discussed in Section $5$, we have found a few instances where distilling an ORE from an $\epsilon$-bounded input was computationally infeasible, thus motivating us to develop and use the kNN-boxes technique for the majority of the results in this paper. In Figure \ref{fig:hard-instances} we compare how OREs grow for increasing values of $\epsilon$ and k (i.e., the control parameters of respectively $\epsilon$-boxes and kNN-boxes).
Finally, in Figure \ref{fig:100inputs-IMDB-OREs} we report a few examples of IMDB reviews that we could solve for an FC with $100$ input words: those examples show OREs for the largest model - in terms of both input size and parameters - that we could solve by means of HS or MSA, eventually improved with the \textit{Adversarial Attacks} routine.

\subsubsection{$\epsilon$-ball Results}
With a perturbation method defined as an $\epsilon$-ball around each input vector (see section \ref{sec:problem-formlation}), Table \ref{tab:mus-vs-hs-1-100} shows a comparison of ORE length and execution time for both the MSA and HS methods.

\begin{figure}
\centering
\includegraphics[height=3.8cm,width=8.3cm]{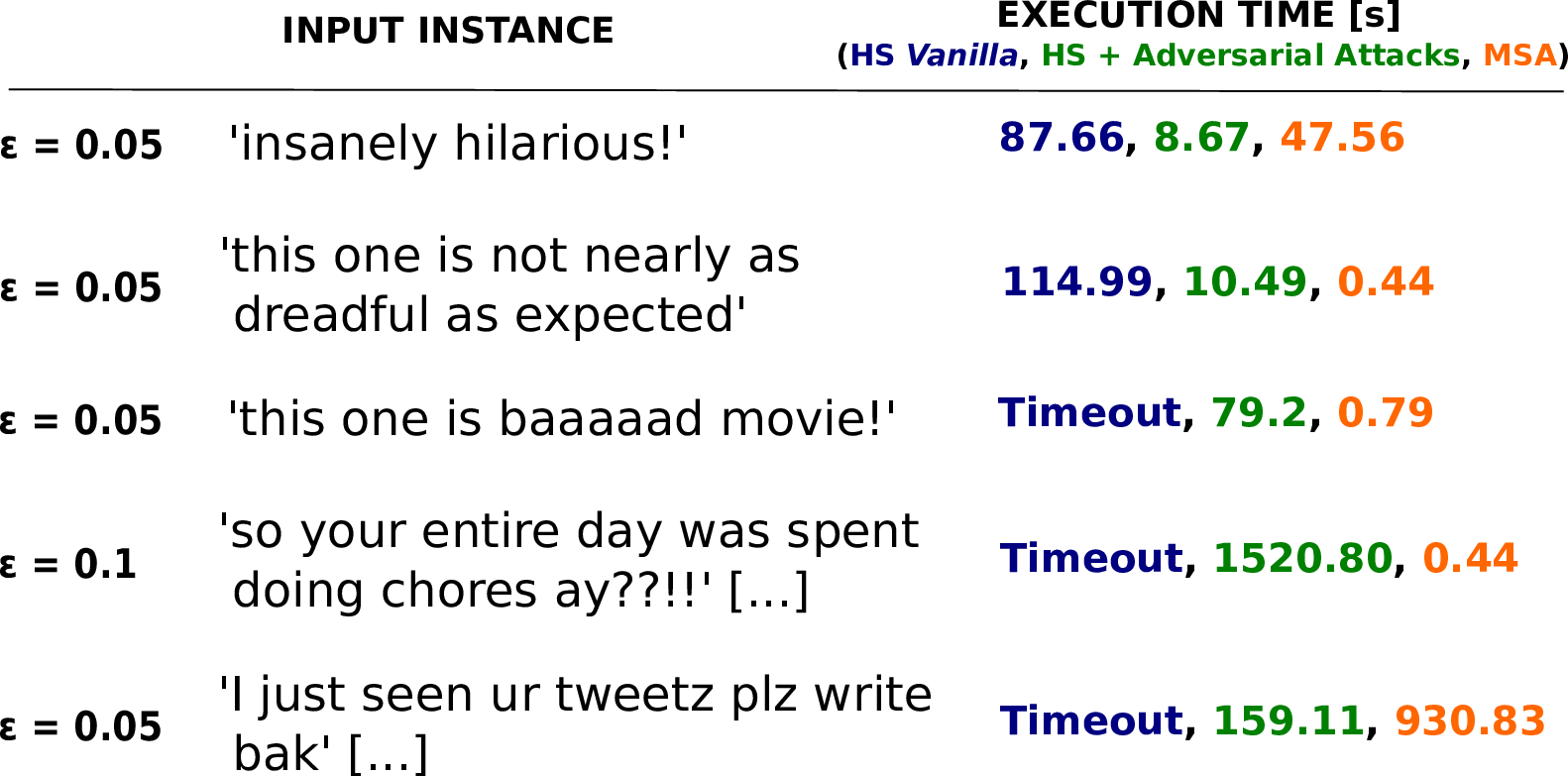}\caption{
Examples of explanations that were enabled by the adversarial attacks routine. Timeout was set to $2$ hours.}
\label{fig:adv-speedup}
\end{figure}

Figure \ref{fig:adv-speedup} shows how using adversarial attacks speeds up convergence.

Below is an example of calculating all possible OREs for a given input and $\epsilon$, and an example of decision bias.

\textbf{Example 1} Calculating \textit{all} of the smallest explanations for an input ($\epsilon=0.05$, FC network, 10 input words, 5 dimensional embedding, SST dataset):

\begin{lstlisting}
Input: ['strange', 'funny', 'twisted', 'brilliant', 'and', 'macabre', '<PAD>', '<PAD>', '<PAD>', '<PAD>']

 Explanations (5 smallest, len=6.0): ['strange', 'funny', 'twisted', 'brilliant', '<PAD>', '<PAD>'] ['strange', 'funny', 'twisted', '<PAD>', '<PAD>', '<PAD>'] ['strange', 'twisted', 'brilliant', '<PAD>', '<PAD>', '<PAD>'] ['strange', 'twisted', 'and', '<PAD>', '<PAD>', '<PAD>'] ['strange', 'twisted', 'macabre', '<PAD>', '<PAD>', '<PAD>']

\end{lstlisting}

\textbf{Example 2} Decision bias, as \texttt{Derrida} cannot be excluded ($\epsilon=0.05$, FC network, 10 input words, 5 dimensional embedding, SST dataset):
\begin{lstlisting}
Input: ['Whether', 'or', 'not', 'you', 'are', 'enlightened', 'by', 'any', 'of', 'Derrida']

Exclude: ['Derrida']

Explanation: ['Whether', 'or', 'are', 'enlightened', 'by', 'any', 'Derrida']

\end{lstlisting}

\clearpage

\end{document}